\documentclass{article}

\usepackage{microtype}
\usepackage{subfigure}
\usepackage{xcolor}
\usepackage{hyperref}
\usepackage{amsmath}
\usepackage{amssymb}
\usepackage{mathtools}
\usepackage{amsthm}
\usepackage[capitalize,noabbrev]{cleveref}
\usepackage{tikz}
\usepackage{placeins}

\usepackage[ruled,vlined,linesnumbered,noend]{algorithm2e}
\usepackage{enumitem}
\SetKwInOut{KwIn}{Input}
\SetKwInOut{KwOut}{Output}
\SetKwInOut{KwInit}{Init}

\usepackage{graphicx}
\usepackage{tikz}
\usetikzlibrary{arrows.meta, positioning, shapes.geometric, fit, backgrounds, calc, matrix}
\newsavebox{\myfigurebox} 
\usepackage{booktabs}
\usepackage{multirow}

\usepackage[preprint]{icml2026}

\icmltitlerunning{DISK: Dynamic Inference SKipping for World Models}

\begin{document}

\twocolumn[
\icmltitle{DISK: Dynamic Inference SKipping for World Models}

\begin{icmlauthorlist}
\icmlauthor{Anugunj Naman}{insta}
\icmlauthor{Gaibo Zhang}{insta}
\icmlauthor{Ayushman Singh}{instb}
\icmlauthor{Yaguang Zhang}{insta}
\end{icmlauthorlist}

\icmlaffiliation{insta}{Purdue University, West Lafayette, United States}
\icmlaffiliation{instb}{Capital One, AI Foundations}

\icmlcorrespondingauthor{Anugunj Naman}{anaman@purdue.edu}

\icmlkeywords{Autoregressive world models, Adaptive inference, Diffusion transformers}

\vskip 0.3in
]

\printAffiliationsAndNotice{} 

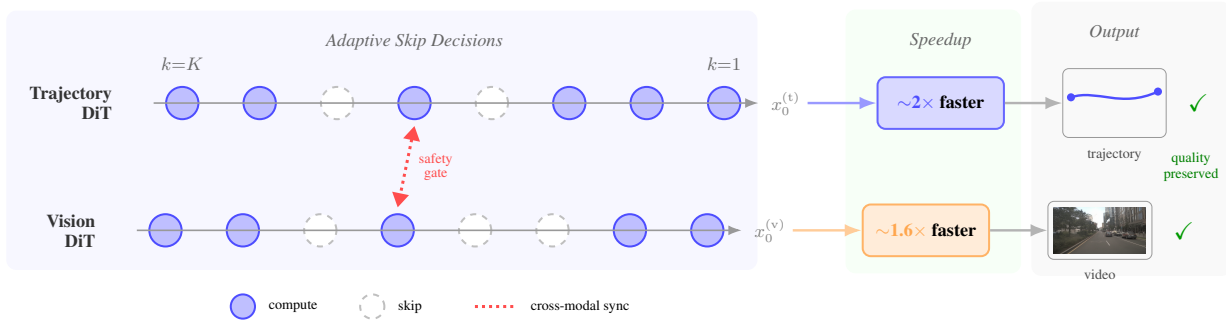
\begin{figure*}[!t]
\centering
\resizebox{0.95\textwidth}{!}{%
\begin{tikzpicture}[
    >=Latex,
    node distance=0.6cm and 0.75cm,
    font=\small,
    compute/.style={circle, draw=blue!70, fill=blue!25, minimum size=16pt, inner sep=0pt, line width=1pt},
    skip/.style={circle, draw=gray!50, dashed, fill=white, minimum size=16pt, inner sep=0pt, line width=1pt},
    branchlabel/.style={font=\small\bfseries, align=right, text=black!80},
    speedbox/.style={draw, rounded corners=4pt, minimum width=2.2cm, minimum height=0.9cm, font=\small\bfseries, line width=1pt},
    outbox/.style={draw=black!40, fill=gray!5, rounded corners=3pt, inner sep=3pt},
    seclabel/.style={font=\footnotesize\itshape, black!60},
    checkstyle/.style={text=green!60!black, font=\large\bfseries},
]


\node[branchlabel] (trajlabel) {Trajectory\\DiT};
\node[branchlabel, below=1.4cm of trajlabel] (vislabel) {Vision\\DiT};

\node[compute, right=0.8cm of trajlabel] (t1) {};
\node[compute, right=of t1] (t2) {};
\node[skip, right=of t2] (t3) {};
\node[compute, right=of t3] (t4) {};
\node[skip, right=of t4] (t5) {};
\node[compute, right=of t5] (t6) {};
\node[compute, right=of t6] (t7) {};
\node[compute, right=of t7] (t8) {};

\node[compute, right=0.8cm of vislabel] (v1) {};
\node[compute, right=of v1] (v2) {};
\node[skip, right=of v2] (v3) {};
\node[compute, right=of v3] (v4) {};
\node[skip, right=of v4] (v5) {};
\node[skip, right=of v5] (v6) {};
\node[compute, right=of v6] (v7) {};
\node[compute, right=of v7] (v8) {};

\draw[->, thick, black!40] (t1.west) ++(-0.2,0) -- ($(t8.east)+(0.3,0)$);
\draw[->, thick, black!40] (v1.west) ++(-0.2,0) -- ($(v8.east)+(0.3,0)$);

\node[font=\footnotesize, black!60, above=0.15cm of t1] {$k{=}K$};
\node[font=\footnotesize, black!60, above=0.15cm of t8] {$k{=}1$};

\node[font=\footnotesize, black!50, right=0.4cm of t8] (toutlabel) {$x_0^{(\mathrm{t})}$};
\node[font=\footnotesize, black!50, right=0.4cm of v8] (voutlabel) {$x_0^{(\mathrm{v})}$};

\draw[red!70, line width=1.5pt, dotted, {Triangle[length=2.5mm]}-{Triangle[length=2.5mm]}] 
    ($(t4.south)+(0,-0.1)$) -- ($(v4.north)+(0,0.1)$);
\node[font=\scriptsize, red!70, align=center, right=0.1cm of $(t4.south)!0.5!(v4.north)$] (gatelabel) {safety\\[-1pt]gate};

\node[seclabel, above=0.5cm of t4] (leftlabel) {Adaptive Skip Decisions};


\node[speedbox, fill=blue!15, draw=blue!60, right=1.2cm of toutlabel] (speedtraj) {\textcolor{blue!70}{$\mathbf{\sim}$2$\times$} faster};
\node[speedbox, fill=orange!15, draw=orange!60, right=1.2cm of voutlabel] (speedvis) {\textcolor{orange!70}{$\mathbf{\sim}$1.6$\times$} faster};

\draw[->, thick, blue!40, line width=1.2pt] (toutlabel.east) -- (speedtraj.west);
\draw[->, thick, orange!40, line width=1.2pt] (voutlabel.east) -- (speedvis.west);

\node[seclabel, above=0.35cm of speedtraj] (centerlabel) {Speedup};


\node[outbox, minimum width=1.8cm, minimum height=1.2cm, right=1.0cm of speedtraj] (outtraj) {};
\node[font=\scriptsize, black!70, below=0.05cm of outtraj.south, anchor=north] {trajectory};

\draw[thick, blue!70, line width=1.2pt] 
    ($(outtraj.west)+(0.15,0.1)$) .. controls ($(outtraj.center)+(-0.3,0.25)$) and ($(outtraj.center)+(0.1,-0.1)$) .. ($(outtraj.east)+(-0.15,0.2)$);
\fill[blue!70] ($(outtraj.west)+(0.15,0.1)$) circle (2pt);
\fill[blue!70] ($(outtraj.east)+(-0.15,0.2)$) circle (2pt);

\node[outbox, right=1.0cm of speedvis] (outvis) {%
    \includegraphics[width=1.6cm, height=1.0cm, keepaspectratio, clip, trim=0 0 0 0]{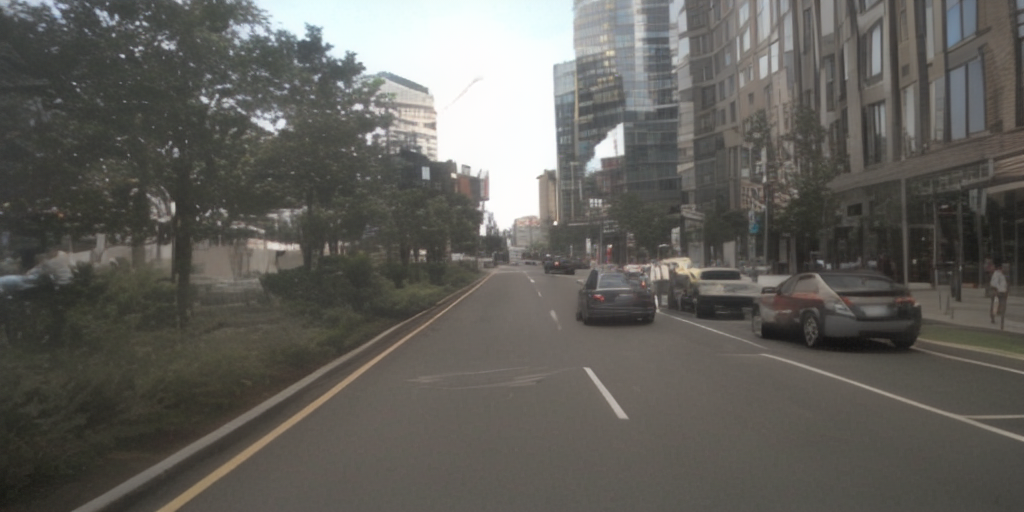}%
};
\node[font=\scriptsize, black!70, below=0.05cm of outvis.south, anchor=north] {video};

\draw[->, thick, black!30, line width=1.2pt] (speedtraj.east) -- (outtraj.west);
\draw[->, thick, black!30, line width=1.2pt] (speedvis.east) -- (outvis.west);

\node[checkstyle, right=0.25cm of outtraj.east] (check1) {$\checkmark$};
\node[checkstyle, right=0.25cm of outvis.east] (check2) {$\checkmark$};

\node[font=\scriptsize, align=center, text=green!50!black] at ($(check1.south)!0.5!(check2.north)$) {quality\\preserved};

\node[seclabel, above=0.35cm of outtraj] (rightlabel) {Output};

\node[compute, minimum size=12pt, below=0.8cm of v2] (legcomp) {};
\node[font=\scriptsize, right=0.1cm of legcomp] (legcomplabel) {compute};

\node[skip, minimum size=12pt, right=0.6cm of legcomplabel] (legskip) {};
\node[font=\scriptsize, right=0.1cm of legskip] (legskiplabel) {skip};

\node[right=0.8cm of legskiplabel] (legline) {};
\draw[red!70, line width=1.5pt, dotted] (legline.west) -- ++(0.7,0);
\node[font=\scriptsize, right=0.85cm of legline.west] {cross-modal sync};

\begin{scope}[on background layer]
    \node[fit=(trajlabel)(vislabel)(t1)(t8)(v1)(v8)(leftlabel)(gatelabel), 
          fill=blue!3, rounded corners=5pt, inner sep=8pt] {};
    \node[fit=(speedtraj)(speedvis)(centerlabel),
          fill=green!3, rounded corners=5pt, inner sep=8pt] {};
    \node[fit=(outtraj)(outvis)(check1)(check2)(rightlabel),
          fill=gray!5, rounded corners=5pt, inner sep=8pt] {};
\end{scope}

\end{tikzpicture}%
}
\caption{\textbf{DISK Overview.} During diffusion sampling, dual-branch controllers decide per-step whether to compute (filled) or skip (dashed) each denoising evaluation. A safety gate synchronizes branches when the trajectory encounters complex maneuvers, ensuring motion-appearance consistency. DISK achieves $\sim$2$\times$ speedup on trajectory diffusion and $\sim$1.6$\times$ on video diffusion while preserving output quality.}
\label{fig:overview}
\end{figure*}

\begin{abstract}
We present DISK, a training-free adaptive inference method for autoregressive world models. DISK coordinates two coupled diffusion transformers—for video and ego-trajectory—via dual-branch controllers with cross-modal skip decisions, preserving motion-appearance consistency without retraining. We extend higher-order latent-difference skip testing to the autoregressive chain-of-forward regime and propagate controller statistics through rollout loops for long-horizon stability. When integrated into closed-loop driving rollouts on 1500 NuPlan and NuScenes samples using an NVIDIA L40S GPU, DISK achieves $\sim$2$\times$ speedup on trajectory diffusion and $\sim$1.6$\times$ on video diffusion while maintaining L2 planning error, visual quality (FID/FVD), and NAVSIM PDMS scores, demonstrating practical long-horizon video-and-trajectory prediction at substantially reduced cost.
\end{abstract}

\section{Introduction}
\label{sec:intro}
World models capture environment dynamics by predicting future observations and agent states, which enables closed-loop decision making and planning~\cite{ha2018wm,lecun2022wm,ding2024wmsurvey}. Diffusion-based transformers (DiT) scale spatiotemporal generation with strong fidelity and controllability~\cite{peebles2023dit, svd, Blattmann2023AlignYL,ma2023efficientdmsurvey}. Recent world model driving systems have used diffusion-based techniques to demonstrate long-horizon rollouts by conditioning on history and forecasting both future scene content and ego motion~\cite{hu2023gaia,gao2024vista,jia2023adriver,wang2023drivedreamer,chen2024drivinggpt,zheng2024doe,zhang2025epona}.

Sampling efficiency is critical because diffusion sampling requires many sequential denoising steps~\cite{ddpm,song2021scorebased}. In autoregressive settings, these costs compound across time as each predicted frame and trajectory conditions the next step. Long rollouts for driving therefore require thousands of model evaluations across two coupled \textit{branches} (vision and trajectory), which stresses latency and energy budgets in real-time applications.

\begin{figure*}[t]
\centering
\begin{tabular}{@{}c@{\hspace{0.5em}}c@{\hspace{0.5em}}c@{\hspace{0.5em}}c@{}}
\small t=0s & \small t=3s & \small t=6s & \small t=9s \\
\includegraphics[width=0.24\textwidth]{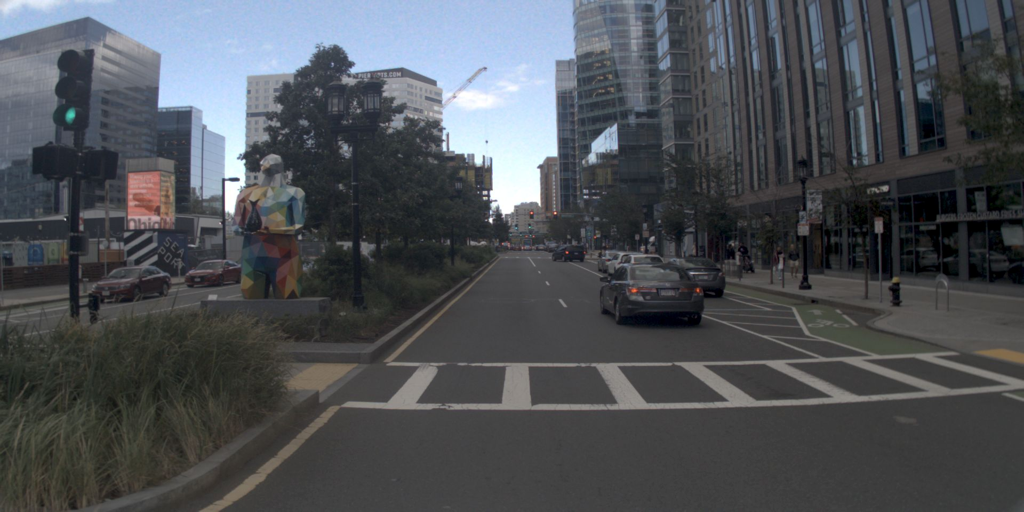} &
\includegraphics[width=0.24\textwidth]{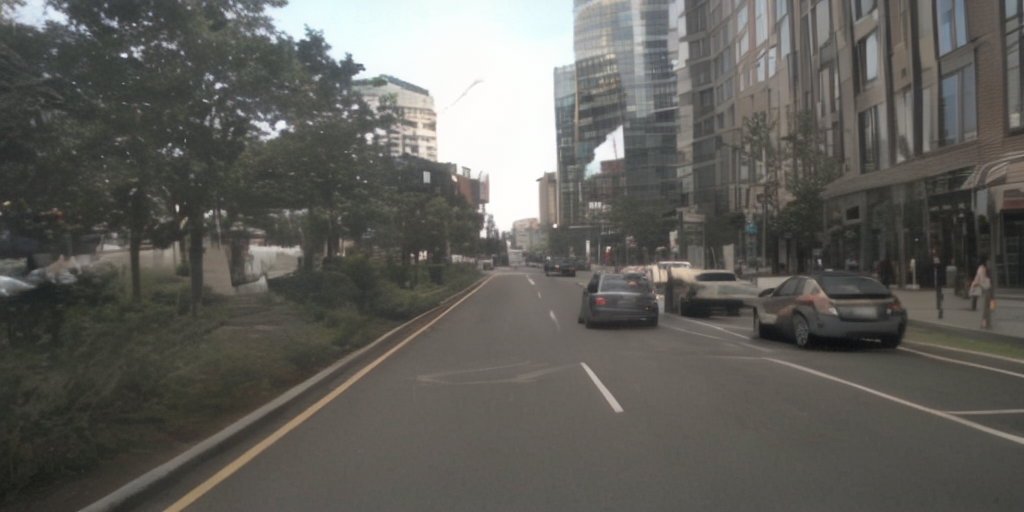} &
\includegraphics[width=0.24\textwidth]{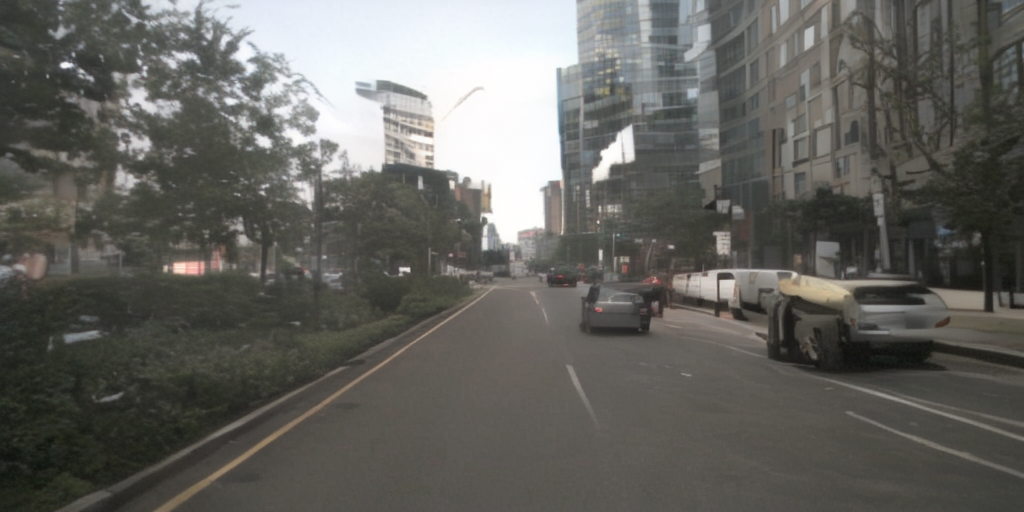} &
\includegraphics[width=0.24\textwidth]{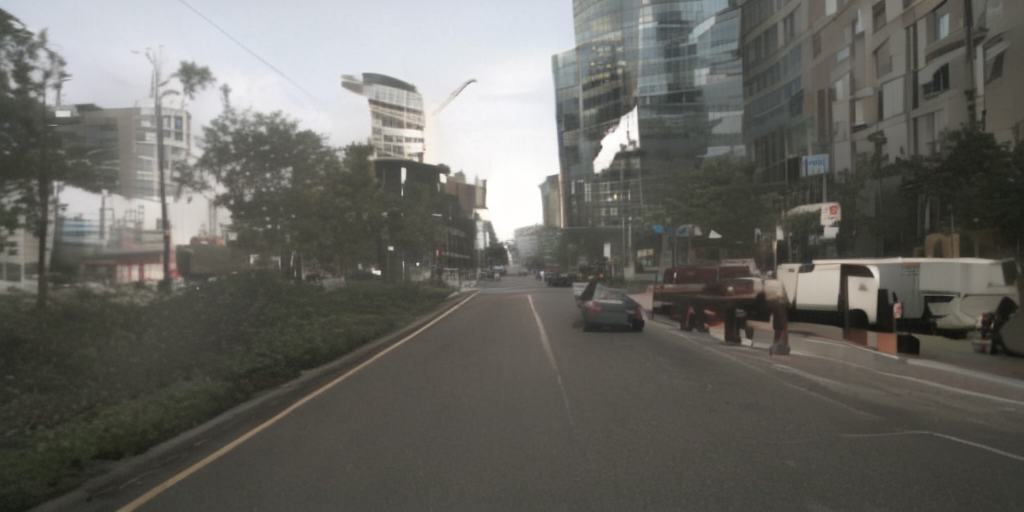} \\
\multicolumn{4}{c}{\small \textit{NuPlan Baseline (Epona)}} \\[0.5em]
\includegraphics[width=0.24\textwidth]{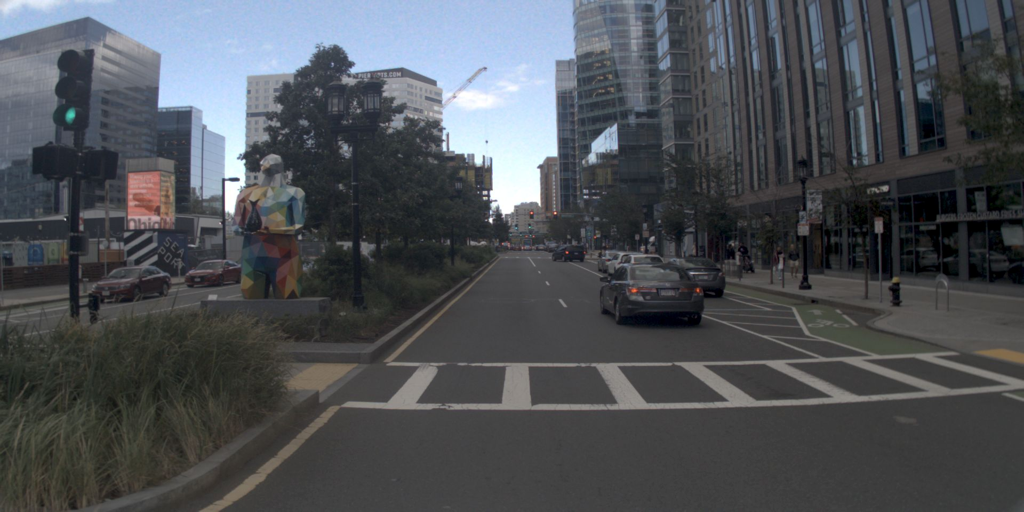} &
\includegraphics[width=0.24\textwidth]{figures/nuplan_disc_3s.png} &
\includegraphics[width=0.24\textwidth]{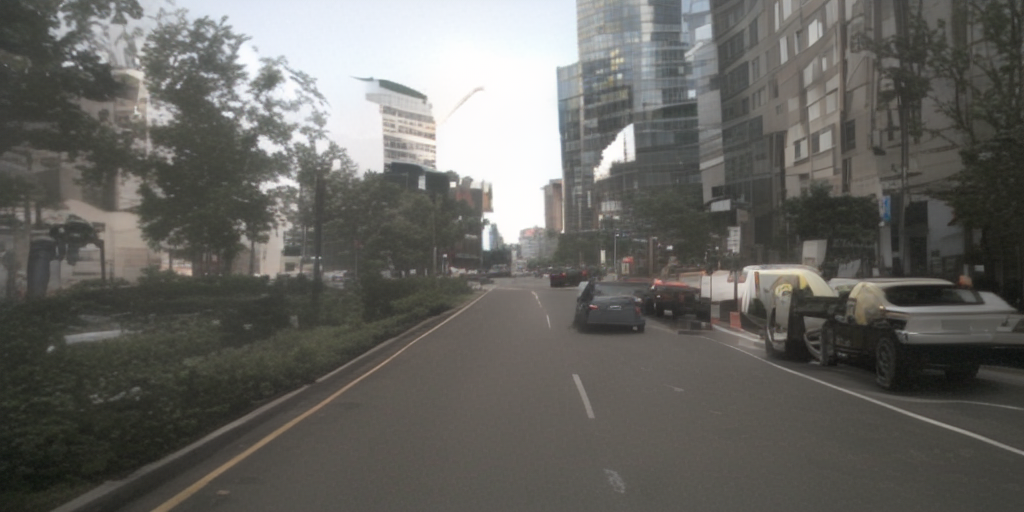} &
\includegraphics[width=0.24\textwidth]{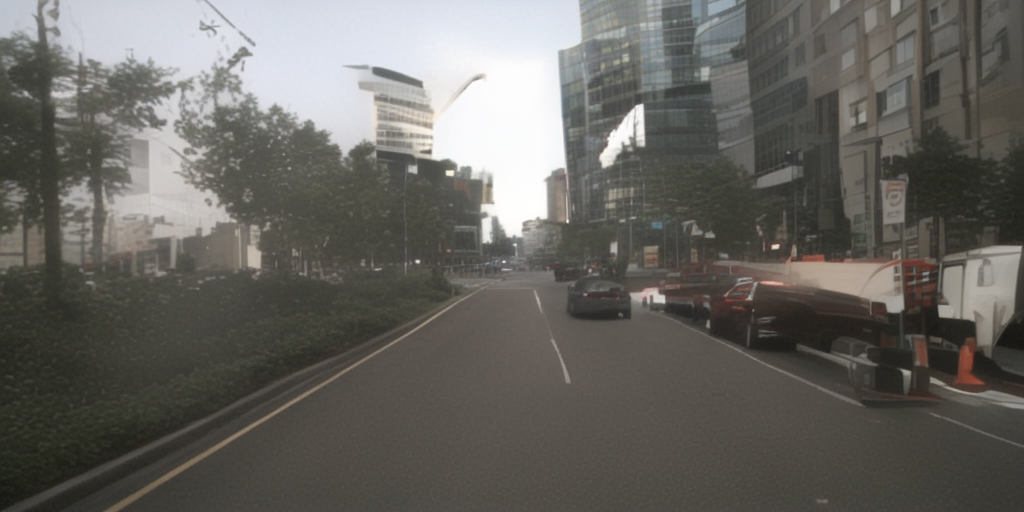} \\
\multicolumn{4}{c}{\small \textit{NuPlan DISK (Ours)}} \\[1em]
\includegraphics[width=0.24\textwidth]{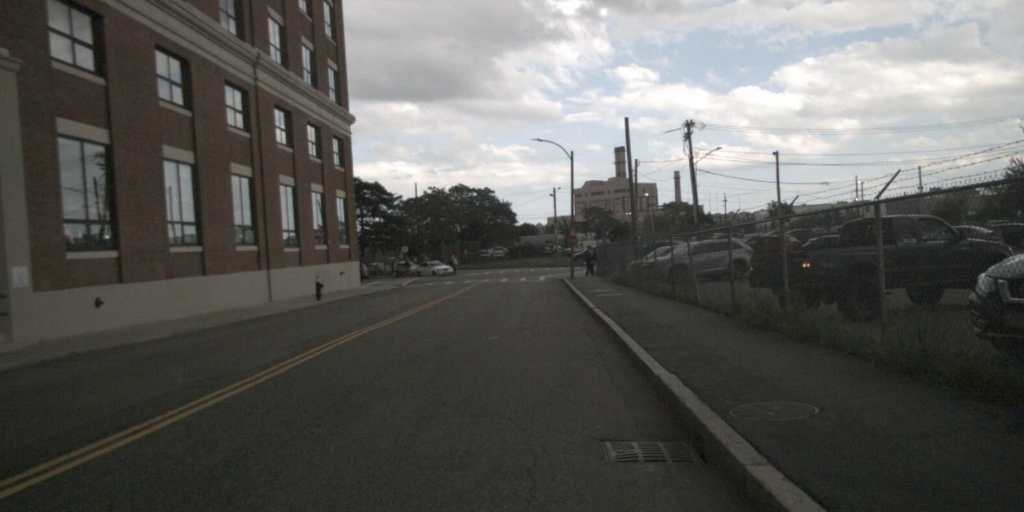} &
\includegraphics[width=0.24\textwidth]{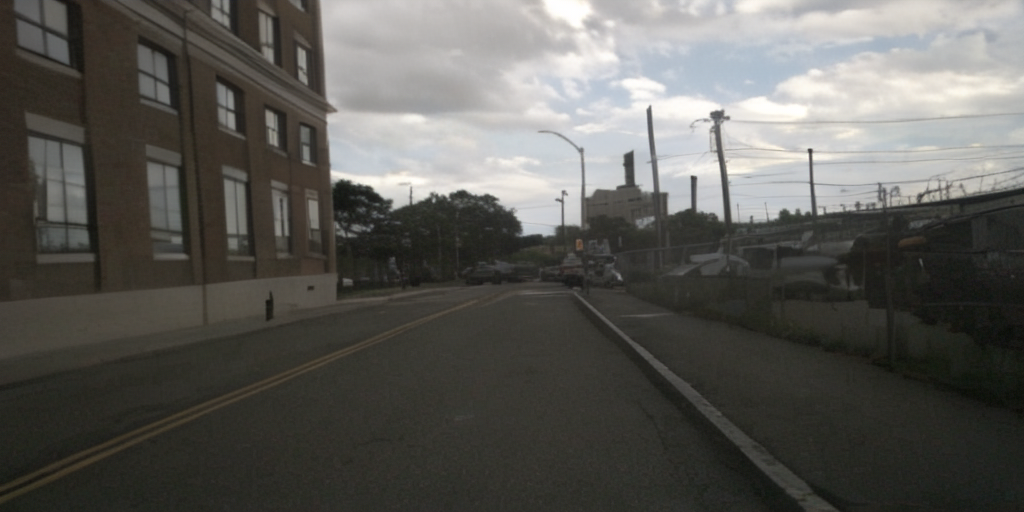} &
\includegraphics[width=0.24\textwidth]{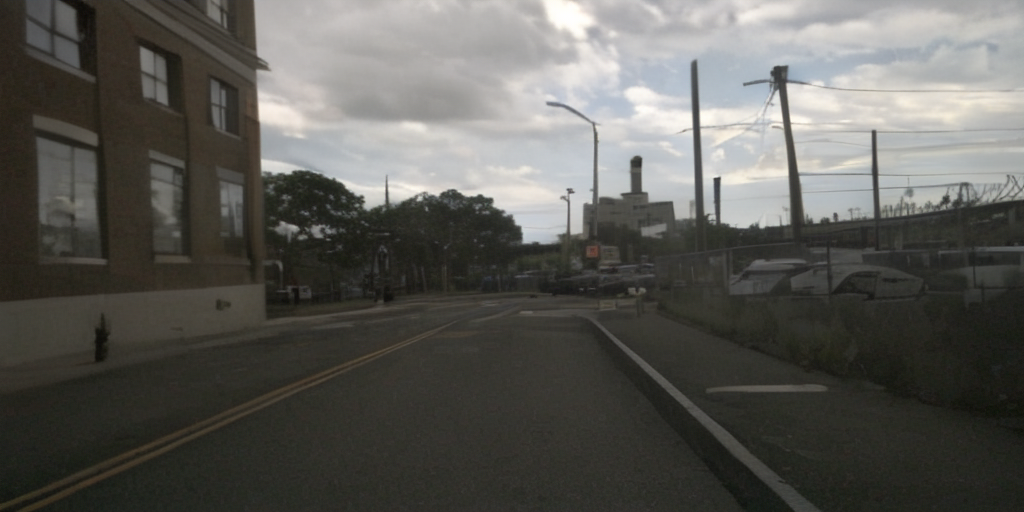} &
\includegraphics[width=0.24\textwidth]{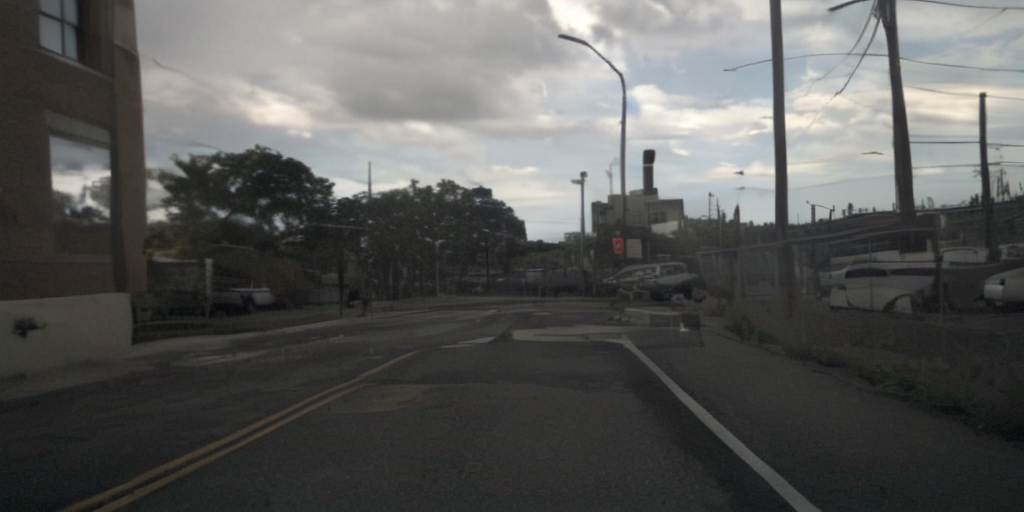} \\
\multicolumn{4}{c}{\small \textit{NuScenes Baseline (Epona)}} \\[0.5em]
\includegraphics[width=0.24\textwidth]{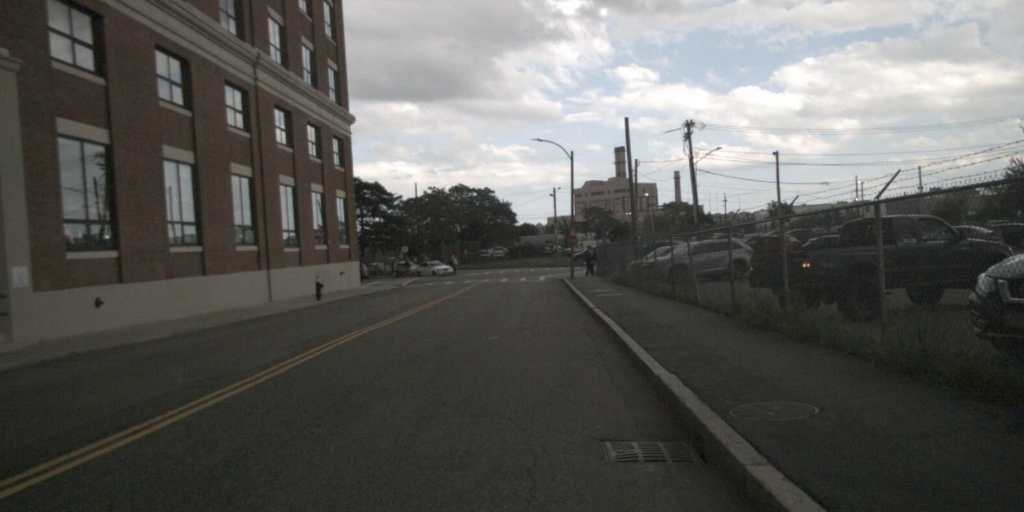} &
\includegraphics[width=0.24\textwidth]{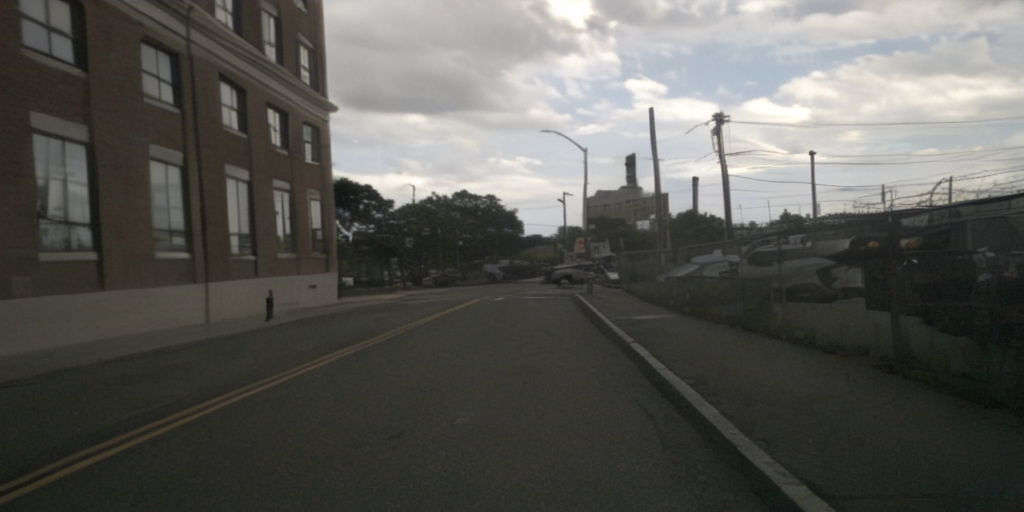} &
\includegraphics[width=0.24\textwidth]{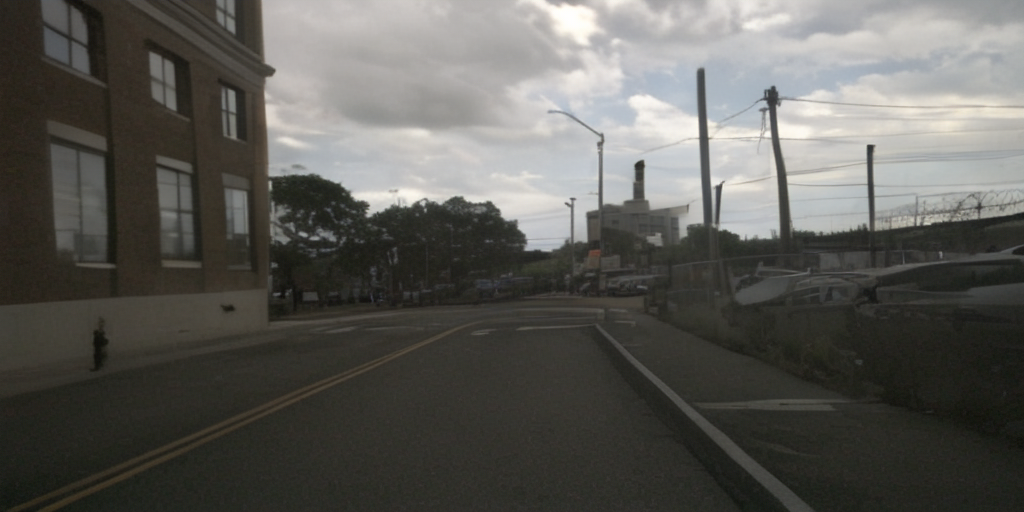} &
\includegraphics[width=0.24\textwidth]{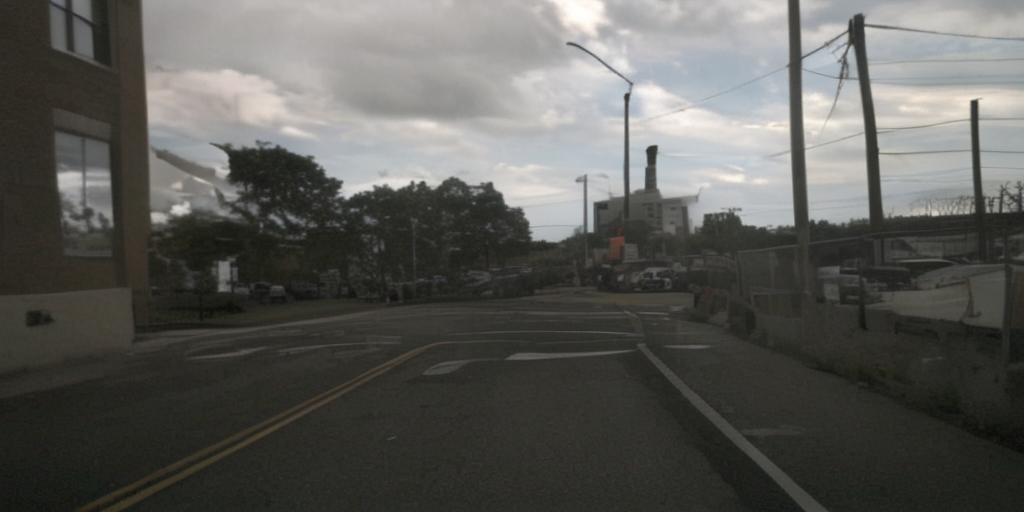} \\
\multicolumn{4}{c}{\small \textit{NuScenes DISK (Ours)}}
\end{tabular}
\caption{\textbf{Qualitative Video Comparison.} Video frames from 11-second rollouts show visually indistinguishable quality between baseline (Epona) and DISK despite $\sim$1.6$\times$ speedup in vision diffusion. DISK maintains high fidelity and temporal consistency across both NuPlan and NuScenes datasets.}
\label{fig:video_comparison}
\end{figure*}

Unfortunately, existing acceleration methods do not fully address this challenge. Fixed-step acceleration or global step reduction degrades fidelity and stability under closed-loop rollout, where small errors accumulate. Training-time accelerators such as consistency models~\cite{song2023consistency,liu2022flow,salimans2021progressive} achieve one-step sampling but require expensive retraining or distillation. This breaks the ``plug-and-play'' flexibility needed to accelerate existing foundation models without altering their learned behavior. Moreover, fixed one-step generation allocates uniform compute regardless of scene complexity. In autoregressive driving rollouts, where static highway cruising alternates with critical maneuvers, this is inefficient: simple frames should be cheap, while complex transitions demand the precision of a full ODE solver to prevent compounding drift. A fixed one-step model cannot adapt to this variation. Inference-time changes to the sampler, such as deterministic formulations or higher-order ODE solvers~\cite{song2020denoising,zhang2022gddim,lu2022dpm,lu2022dpm++}, do not adapt to per-step difficulty. Furthermore, they fail to coordinate the visual and trajectory predictions in autoregressive world models. Other training-free accelerators, including early exiting and feature caching~\cite{tang2023deediff,ma2023deepcache}, or automated step selection~\cite{li2023autodiffusion,redi2024}, optimize computation within a single denoising network but do not coordinate cross-modal branches in a closed-loop world-model rollout.

In contrast, we propose Dynamic Inference SKipping (DISK), a training-free adaptive inference module that operates purely at test time. DISK equips the vision and trajectory DiTs with dual skip controllers and introduces cross-modal coordination that keeps the two branches synchronized. An adaptive per-step criterion governs compute-or-skip decisions, while controller statistics propagate through the autoregressive chain to stabilize long-horizon rollouts. This design preserves model weights and training procedures, drops into existing world-model pipelines, and explicitly targets the challenges of closed-loop driving under chain-of-forward~\cite{zhang2025epona} self-conditioning.

Our controller maintains short histories of latent states for each branch to estimate local curvature and step difficulty. The vision and trajectory controllers exchange lightweight summaries to avoid divergent skipping patterns that could desynchronize appearance from motion. We instrument the rollout loop to log skip dynamics and analyze when and where compute is most needed, which provides insight into the interplay between autoregressive drift and denoising difficulty.

We integrate DISK into long-horizon driving rollouts and analyze skip dynamics, compute/skip ratios, and their effect on both visual quality and planning accuracy. Evaluated on 1500 samples from both NuPlan~\cite{caesar2021nuplan} and NuScenes~\cite{caesar2020nuscenes} on an NVIDIA L40S GPU, DISK achieves $\sim$2$\times$ speedup on the trajectory branch (TrajDiT: 20ms $\to$ 10ms) and $\sim$1.6$\times$ on the vision branch (VisDiT: 244ms $\to$ 155ms), while preserving L2 planning error, visual quality (FID 12.5, FVD 88.1), collision rates, and NAVSIM PDMS scores. These results demonstrate that training-free adaptive inference substantially reduces cost without compromising downstream control.

We summarize our contributions as follows:
\begin{itemize}[itemsep=0pt, parsep=0pt, topsep=0pt, partopsep=0pt, leftmargin=*]
    \item \textbf{A training-free adaptive inference framework} that orchestrates visual and trajectory diffusion transformers. By employing adaptive per-step skip decisions, it dynamically reduces computational cost without requiring retraining.
    \item \textbf{A cross-modal skip coordination mechanism} that harmonizes the visual and trajectory branches. This mechanism propagates controller statistics through the autoregressive loop to ensure synchronization and preserve long-horizon temporal consistency.
\end{itemize}

Extensive empirical evaluation on NuPlan~\cite{caesar2021nuplan} and NuScenes~\cite{caesar2020nuscenes} demonstrates that our method achieves significant inference acceleration while preserving planning accuracy, visual fidelity, and stability (see Figure~\ref{fig:video_comparison}).

\section{Related Work}
\label{sec:related}

\textbf{Autoregressive World Models.}
World models capture environment dynamics to enable planning and control. Recent approaches leverage diffusion transformers for high-fidelity spatiotemporal generation~\cite{hu2023gaia,gao2024vista,jia2023adriver,wang2023drivedreamer,chen2024drivinggpt,zheng2024doe,zhang2025epona}. Beyond driving, researchers have advanced world models via discrete latents~\cite{hafner2020discrete,hafner2023mastering}, sample-efficient transformer architectures~\cite{micheli2023transformers}, and efficient stochastic transformers~\cite{zhang2023storm}. See~\cite{ding2024wmsurvey} for a survey. While these systems differ in representation, they share a closed-loop evaluation regime where errors accumulate across long horizons. Our work targets inference efficiency for such autoregressive pipelines while preserving rollout fidelity and planning accuracy.

\textbf{Efficient Diffusion Sampling.}
Sampling in diffusion models requires many sequential denoising steps~\cite{ddpm,song2021scorebased,ma2023efficientdmsurvey}. Deterministic samplers and higher-order ODE solvers reduce steps without retraining, but they do not modify the underlying score model and are therefore limited by the higher-order compute inversely related to number of steps~\cite{song2020denoising,zhang2022gddim,lu2022dpm,lu2022dpm++,zheng2023dpm,ma2023efficientdmsurvey}. Training-time accelerators, such as progressive distillation and consistency models, learn faster sampling but require additional compute and may introduce distribution shift~\cite{salimans2021progressive,song2023consistency}. System-level methods exploit redundancy through early exiting and feature caching~\cite{tang2023deediff,ma2023deepcache,lyu2022accelerating}, automated schedule search~\cite{li2023autodiffusion}, or parallel denoising across timesteps~\cite{chen2025hvad}. These methods typically treat a single noise-prediction network, and \textit{none} address coordination between coupled branches in autoregressive world models.

\textbf{Adaptive Inference.}
Test-time adaptation allocates computation where it matters without changing model weights. Approaches span schedule design, region-adaptive sampling~\cite{tian2025bottleneck,liu2025region}, and system-level sparsity~\cite{fang2024xdit,zhang2025jenga,xia2025sparsevideo,zhang2025spargeattn}. One line of work estimates per-step difficulty and skips denoising evaluations to save forward passes~\cite{adaptivediffusion24ye}. Prior work generally targets a single UNet or DiT branch in standard AIGC settings. We study adaptive inference for dual-branch autoregressive world models with coupled vision and trajectory, where cross-modal coordination and long-horizon stability are central concerns.

\textbf{Summary of Differences.}
In contrast to sampler redesign or training-time distillation, DISK is training-free and focuses on per-step adaptivity. Unlike early exiting or caching that optimize a single branch, we coordinate skip decisions between vision and trajectory branches and propagate controller statistics across the autoregressive loop. Concretely, we:
\begin{itemize}[itemsep=0pt, parsep=0pt, topsep=0pt, partopsep=0pt, leftmargin=*]
  \item Target an autoregressive world model with two coupled DiTs (vision and trajectory);
  \item Introduce dual-branch controllers with cross-modal coordination for appearance-motion consistency;
  \item Preserve chain-of-forward self-conditioning while skipping;
  \item Evaluate on closed-loop driving with planning metrics and analyze skip dynamics across time.
\end{itemize}
This enables substantial speedup on both trajectory and video diffusion while preserving planning quality and visual fidelity.

\section{Method}
\label{sec:method}

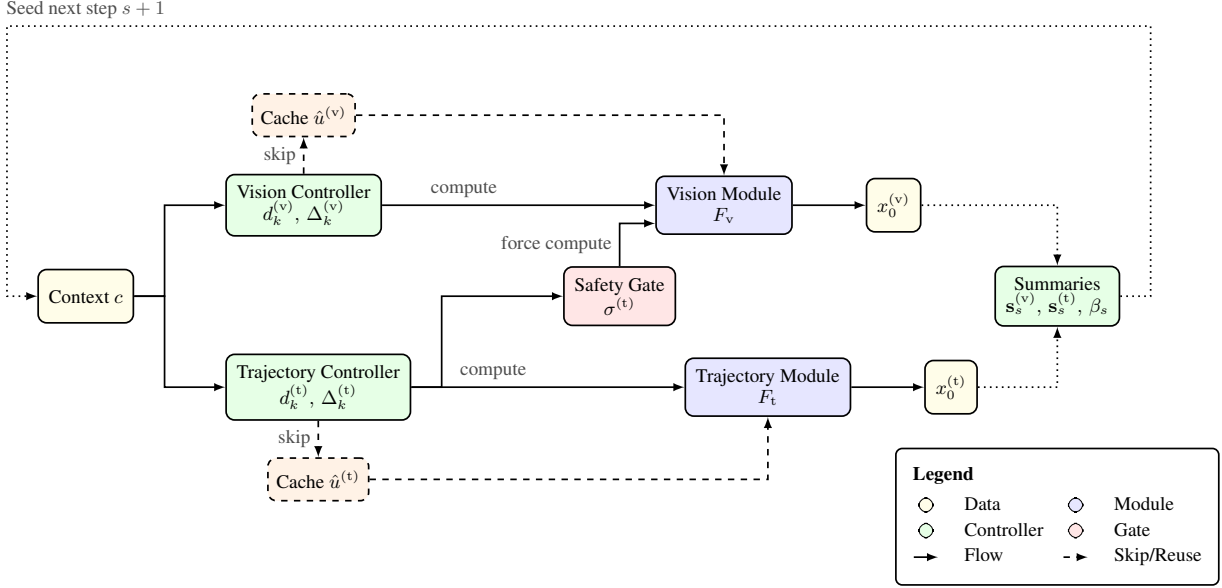
\begin{figure*}[t]
\centering

\sbox{\myfigurebox}{%
\begin{tikzpicture}[
    node distance=1.5cm and 2.0cm,
    >=latex, thick,
    font=\small,
    data/.style={draw, rounded corners, fill=yellow!10, align=center, minimum height=2.5em, inner sep=5pt},
    process/.style={draw, rounded corners, fill=blue!10, align=center, minimum height=2.5em, inner sep=5pt},
    ctrl/.style={draw, rounded corners, fill=green!10, align=center, minimum height=2.5em, inner sep=5pt},
    gate/.style={draw, rounded corners, fill=red!10, align=center, minimum height=2.5em, inner sep=5pt},
    cache/.style={draw, dashed, rounded corners, fill=orange!10, align=center, minimum height=2em, inner sep=4pt},
    note/.style={font=\footnotesize, color=black!70}
]

  \node[data] (ctx) {Context $c$};

  \node[ctrl, right=1.5cm of ctx, yshift=1.5cm] (vctrl) {Vision Controller\\$d_k^{(\mathrm{v})},\,\Delta_k^{(\mathrm{v})}$};
  \node[ctrl, right=1.5cm of ctx, yshift=-1.5cm] (tctrl) {Trajectory Controller\\$d_k^{(\mathrm{t})},\,\Delta_k^{(\mathrm{t})}$};

  \node[gate, right=2.5cm of tctrl, yshift=1.5cm] (gate) {Safety Gate\\$\sigma^{(\mathrm{t})}$};

  \node[process, right=4.5cm of vctrl] (vnet) {Vision Module\\$F_{\mathrm{v}}$};
  \node[process, right=4.5cm of tctrl] (tnet) {Trajectory Module\\$F_{\mathrm{t}}$};

  \node[cache, above=0.6cm of vctrl] (vcache) {Cache $\hat{u}^{(\mathrm{v})}$};
  \node[cache, below=0.6cm of tctrl] (tcache) {Cache $\hat{u}^{(\mathrm{t})}$};

  \node[data, right=1.2cm of vnet] (vout) {$x_0^{(\mathrm{v})}$};
  \node[data, right=1.2cm of tnet] (tout) {$x_0^{(\mathrm{t})}$};

  \node[ctrl, right=1.2cm of vout, yshift=-1.5cm] (sum) {Summaries\\$\mathbf{s}_s^{(\mathrm{v})},\,\mathbf{s}_s^{(\mathrm{t})},\,\beta_s$};

  
  \draw[->] (ctx.east) -- ++(0.5,0) |- (vctrl.west);
  \draw[->] (ctx.east) -- ++(0.5,0) |- (tctrl.west);

  \draw[->] (vctrl.east) -- node[above, note, pos=0.3] {compute} (vnet.west);
  \draw[->] (tctrl.east) -- node[above, note, pos=0.3] {compute} (tnet.west);

  \draw[->] (tctrl.east) -- ++(0.5,0) |- (gate.west);

  \draw[->] (gate.north) |- node[left, note, pos=0.25] {force compute} ([yshift=-0.3cm]vnet.west);

  \draw[->, dashed] (vctrl.north) -- node[left, note] {skip} (vcache.south);
  \draw[->, dashed] (tctrl.south) -- node[left, note] {skip} (tcache.north);

  \draw[->, dashed] (vcache.east) -| (vnet.north);
  \draw[->, dashed] (tcache.east) -| (tnet.south);

  \draw[->] (vnet.east) -- (vout.west);
  \draw[->] (tnet.east) -- (tout.west);

  \draw[->, dotted] (vout.east) -| (sum.north);
  \draw[->, dotted] (tout.east) -| (sum.south);
  \draw[->, dotted] (sum.east) -- ++(0.5,0) |- ([yshift=4.0cm]ctx.north) -| ([xshift=-0.5cm]ctx.west) -- (ctx.west);
  \node[note, above, align=center] at ([yshift=4.0cm]ctx.north) {Seed next step $s+1$};


\node[draw, rounded corners=3pt, fill=white, 
      below=2cm of sum, inner sep=8pt] (legend) {
  \begin{tabular}{@{}cl@{\quad}cl@{}}
    \multicolumn{4}{@{}l@{}}{\textbf{Legend}} \\[4pt]
    \tikz[baseline=-1.5ex]\node[data, minimum size=0.75em, inner sep=1pt] {}; & Data &
    \tikz[baseline=-1.5ex]\node[process, minimum size=0.75em, inner sep=1pt] {}; & Module \\[2pt]
    \tikz[baseline=-1.5ex]\node[ctrl, minimum size=0.75em, inner sep=1pt] {}; & Controller &
    \tikz[baseline=-1.5ex]\node[gate, minimum size=0.75em, inner sep=1pt] {}; & Gate \\[2pt]
    \tikz[baseline=-0.5ex]\draw[->, thick] (0,0) -- (0.4,0); & Flow &
    \tikz[baseline=-0.5ex]\draw[->, dashed, thick] (0,0) -- (0.4,0); & Skip/Reuse
  \end{tabular}
};

\end{tikzpicture}
}

\resizebox{0.95\textwidth}{!}{\usebox{\myfigurebox}}

\caption{Overview of DISK inference. The Trajectory Controller influences the Vision branch via a unidirectional Safety Gate, ensuring visual consistency during complex maneuvers. The Vision Controller operates independently but is overridden if the Safety Gate triggers. Statistics from both branches seed the next autoregressive step.}
\label{fig:adaptive_overview}
\end{figure*}

We formulate the problem (§3.1), describe the dual-controller architecture (§3.2), and analyze the safety gate (§3.3). We introduce a training-free adaptive inference module for an autoregressive world model with two coupled diffusion transformers: a visual branch that predicts the next video latent and a trajectory branch that predicts the future ego motion. The core idea is to decide, at each diffusion step and for each branch, whether to compute a fresh prediction or reuse a cached one, and to coordinate these decisions so appearance and motion remain consistent over long horizons. Figure~\ref{fig:adaptive_overview} illustrates the overall architecture, showing how context flows through controllers to modules, with the safety gate coordinating cross-modal decisions.

\subsection{Model and Notation}
\label{subsec:problem}
We index autoregressive rollout steps by $s$. At each step, the model predicts a visual latent and an ego-trajectory latent conditioned on a context $c$ (past frames, poses, and other features). Both branches are sampled on a descending diffusion time grid $\{\tau_k\}_{k=0}^{K}$ with $\tau_0<\tau_1<\cdots<\tau_K$ and step index $k\in\{K,\ldots,1\}$. We denote by $x_k^{(b)}$ the latent at diffusion step $k$ for branch $b\in\{\mathrm{v},\mathrm{t}\}$, and by $F_b(\cdot;\tau,c)$ the branch network that outputs a velocity/noise prediction. A generic sampler update is
\begin{equation}
\label{eq:update}
x_{k-1}^{(b)}\;=\;\Psi\bigl(x_k^{(b)},\;\tau_k,\;F_b(x_k^{(b)};\tau_k,c)\bigr)\, ,\quad k=1,\dots,K,
\end{equation}
where $\Psi$ is the chosen numerical update (Euler or an ODE solver). Our method leaves $F_b$ and $\Psi$ unchanged.

\paragraph{Notation at a glance.}
\begin{itemize}[leftmargin=*]
  \item $s$: rollout step in the autoregressive loop; $k$: diffusion step on the grid $\{\tau_k\}_{k=0}^{K}$; $K$: number of diffusion steps.
  \item $b\in\{\mathrm{v},\mathrm{t}\}$: branch indicator (visual or trajectory).
  \item $x_k^{(b)}$: branch latent at diffusion step $k$; $x_0^{(b)}$ is the final latent used downstream (decoding a frame for $\mathrm{v}$, reading poses/yaws for $\mathrm{t}$).
  \item $F_b(\cdot;\tau,c)$: branch network evaluated at time $\tau$ and context $c$; $\Psi$ : sampler update in Eq.~\eqref{eq:update}.
  \item $d_k^{(b)}=\operatorname{mean}\,\lvert x_k^{(b)}{-}x_{k-1}^{(b)}\rvert$: scalar difficulty proxy.
  \item $\Delta_k^{(b)}=\bigl\lvert\tfrac{1}{2}(d_k^{(b)}{+}d_{k+2}^{(b)})-d_{k+1}^{(b)}\bigr\rvert$: local linearity/curvature check (reverse-time window).
  \item $\theta$: relative tolerance controlling skip aggressiveness; $W$: warm-up steps; $C_{\max}$: maximum consecutive skips; $\varepsilon$: stall threshold.
  \item $m_k^{(b)}\in\{\texttt{compute},\texttt{skip}\}$: decision at step $k$; $\text{consecutive\_skips}^{(b)}$: running counter per branch.
  \item $\sigma_k^{(b)}\in\{0,1\}$: safety gate for step $k$ (computed from the latest available $d$); $\hat{u}_k^{(b)}$: cached branch prediction when skipping.
  \item $\mathbf{s}_s^{(b)}$: per-branch summary passed to step $s{+}1$ (compute/skip counts and safety fraction $\rho_s^{(b)}$).
\end{itemize}

\subsection{Adaptive Per-Step Skipping}
\label{subsec:method}
We maintain a short history of latents and decide at each $k$ whether to recompute $F_b$ or reuse the previous prediction. The controller uses simple, scale-free statistics and has three components: a local smoothness test, caching, and safety guards.

\noindent\textbf{Local smoothness.} For branch $b$, we define the mean absolute change as the difficulty proxy (diff):
\begin{equation}
\label{eq:diff}
d_k^{(b)}\;=\;\operatorname{mean}\,\lvert x_k^{(b)}-x_{k-1}^{(b)}\rvert\, .
\end{equation}
When iterating from $k{=}K$ down to $1$, the most recent three diffs in the FIFO buffer correspond to $\{d_{k+2}^{(b)},d_{k+1}^{(b)},d_k^{(b)}\}$. We test whether the local trend is nearly linear via
\begin{equation}
\label{eq:curv}
\Delta_k^{(b)}\;=\;\Bigl\lvert\tfrac{1}{2}\bigl(d_k^{(b)}+d_{k+2}^{(b)}\bigr)-d_{k+1}^{(b)}\Bigr\rvert\, .
\end{equation}
We compute $d_k^{(b)}$ after forming $x_{k-1}^{(b)}$ at step $k$, so $\Delta_k^{(b)}$ is available at the end of step $k$. We use this just-computed $\Delta_k^{(b)}$ to set the \emph{next-step} decision $m_{k-1}^{(b)}$: we skip if $\Delta_k^{(b)}\le \theta\,d_{k+1}^{(b)}$ and compute otherwise, where $\theta>0$ tunes aggressiveness. Until the three-diff buffer is filled, we default to \texttt{compute}. The goal is to detect stability in the evolution of latent changes over time using a \textit{second-order finite-difference criterion}. Here $\Delta_k^{(b)}$ is the second-order finite-difference residual on $\{d_{k+2}^{(b)}, d_{k+1}^{(b)}, d_k^{(b)}\}$; it is zero when $d_j^{(b)}$ is locally affine in $j$. The controller skips predictable low-$\Delta_k^{(b)}$ regions.


\noindent\textbf{Caching and reuse.} On a compute step we cache $F_b(x_k^{(b)};\tau_k,c)$. On a skip we reuse the cache inside $\Psi$ in Eq.~\eqref{eq:update}, preserving the time grid while avoiding a forward pass.

\noindent\textbf{Safety guards.} The safety guards consist of three different checks. We use a warm-up of $W$ initial steps with mandatory compute, a cap of $C_{\max}$ consecutive skips, and a stall check that forces compute when the most recent diff $d_{k+1}^{(b)}$ is numerically tiny. The stall threshold $\varepsilon$ is a \emph{guard} that prevents degenerate skipping near zero change, while the curvature tolerance $\theta$ is a \emph{smoothness} threshold used in the local test ($\Delta_k^{(b)}\le \theta\,d_{k+1}^{(b)}$). In practice we set $\varepsilon\approx 10^{-6}$ to match implementation, but values on the order of machine epsilon are also acceptable. 

\begin{figure*}[t]
\centering
\begin{tikzpicture}[scale=0.8, every node/.style={font=\small}]
  \draw[->, thick, gray] (0,0) -- (11.5,0) node[right, black] {Diffusion steps $k$};
  \foreach \x in {0,2,4,6,8,10}
    \draw[gray] (\x,0.08) -- (\x,-0.08) node[below, black] {\x};
  
  \draw[blue!70, line width=1.5pt] (0,1) -- (11,1);
  \node[left, blue!80!black, font=\small\bfseries] at (-0.1,1) {Vision};
  \foreach \x in {0,1,3,4,6,9,10} {
    \fill[blue] (\x,1) circle (2.5pt);
  }
  \foreach \x in {2,5,7,8} {
    \draw[blue!60, thick] (\x-0.15,0.85) -- (\x+0.15,1.15);
    \draw[blue!60, thick] (\x+0.15,0.85) -- (\x-0.15,1.15);
  }
  \node[blue!70!black, right, font=\footnotesize] at (11.1,1) {Compute: 7, Skip: 4};
  
  \draw[red!70, line width=1.5pt] (0,2) -- (11,2);
  \node[left, red!80!black, font=\small\bfseries] at (-0.1,2) {Trajectory};
  \foreach \x in {0,1,2,4,6,8,9,10} {
    \fill[red] (\x,2) circle (2.5pt);
  }
  \foreach \x in {3,5,7} {
    \draw[red!60, thick] (\x-0.15,1.85) -- (\x+0.15,2.15);
    \draw[red!60, thick] (\x+0.15,1.85) -- (\x-0.15,2.15);
  }
  \node[red!70!black, right, font=\footnotesize] at (11.1,2) {Compute: 8, Skip: 3};
  
  \draw[red!40, densely dotted, line width=1.2pt] (4,1.92) -- (4,1.08);
  \draw[red!40, densely dotted, line width=1.2pt] (9,1.92) -- (9,1.08);
  \node[red!50!black, above, font=\scriptsize\itshape] at (4,2.15) {Safety};
  \node[red!50!black, above, font=\scriptsize\itshape] at (9,2.15) {Safety};
  
  \node[draw=gray!50, rounded corners=2pt, fill=white, anchor=north, inner sep=6pt] at (5.5,-0.8) {
    \tikz[baseline=-0.65ex]\fill[blue] (0,0) circle (2pt); Compute step \quad
    \tikz[baseline=-0.65ex]\draw[blue!60, thick] (-0.12,-0.12) -- (0.12,0.12) (-0.12,0.12) -- (0.12,-0.12); Skip step \quad
    \tikz[baseline=-0.5ex]\draw[red!40, densely dotted, line width=1.2pt] (0,0) -- (0.4,0); Safety gate trigger
  };
\end{tikzpicture}
\caption{\textbf{Adaptive Skipping Pattern.} Example skip decisions across 11 diffusion steps for vision and trajectory branches. Filled circles indicate compute steps, crosses indicate skips. The safety gate triggers (dotted lines) force vision compute when trajectory encounters complex maneuvers.}
\label{fig:skipping_mechanism}
\end{figure*}
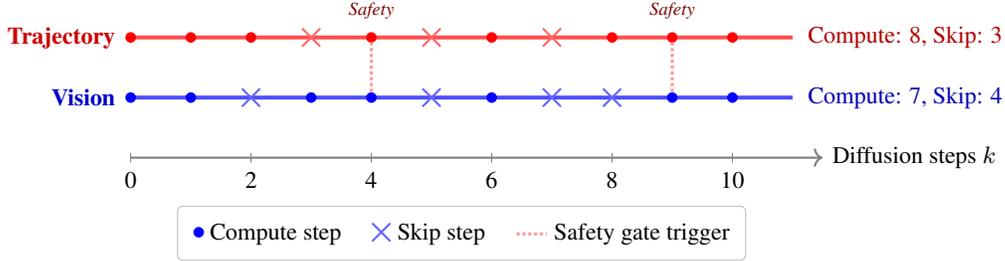

\noindent\textbf{Practical settings.} Unless stated otherwise the default settings are $\theta{=}0.01$, $W{=}3$, and $C_{\max}{=}4$. Figure~\ref{fig:skipping_mechanism} shows an example of the resulting skip pattern across diffusion steps, illustrating how the controllers adaptively decide when to compute or skip based on local smoothness, and how the safety gate coordinates the two branches.

\subsection{Coordinating the Two Branches}
\label{subsec:gate}
Independently skipping vision and trajectory can desynchronize appearance and motion. We add light-weight coupling that make the policy robust under closed-loop rollouts.

\noindent\textbf{Trajectory safety signal.} Let $\sigma_k^{(t)}\in\{0,1\}$ indicate whether \textit{any} of the three safety guards is active for the trajectory branch at step $k$:
\begin{equation}
\begin{aligned}
\sigma_k^{(t)} = &\; \mathbf{1}\{k > K{-}W\} \\
&\;\vee\; \mathbf{1}\{\text{consecutive\_skips}^{(t)} = C_{\max}\} \\
&\;\vee\; \mathbf{1}\{d_{k+1}^{(t)} \le \varepsilon\}\, .
\end{aligned}
\end{equation}
In the reverse-time loop, $d_{k+1}^{(t)}$ is the most recent diff available when deciding step $k$; if the buffer is empty (early steps), the stall term is skipped.

\noindent\textbf{Unidirectional coupling.} To prioritize throughput we keep the original branch decisions and only force compute on the vision branch when the trajectory safety signal triggers:
\begin{equation}
    \label{eq:async_gate}
    \text{if } \sigma_k^{(\mathrm{t})}{=}1 \text{ then } m_k^{(\mathrm{v})}\leftarrow \texttt{compute}
\end{equation}
This unidirectional coupling ensures that if the ego-vehicle executes a complex maneuver (triggering trajectory compute), the video model also computes to render the corresponding visual changes. However, visual complexity (triggering vision compute) does not force trajectory compute, saving time on the slower video branch when trajectory is simple.
Trajectory safety is unaffected by vision, so we only run one pass per branch. The gate does not affect the safety ratio $\rho_s^{(\mathrm{v})}$; $\rho_s$ tracks each branch’s local safety triggers (warm-up, consecutive-skip cap, stall) and excludes gate-forced computes. This mirrors the behavior of the original Epona~\cite{zhang2025epona} sampler and yields the $\sim 2\times$ acceleration reported in Sec.~\ref{sec:exp}, at the cost of weaker appearance–motion coupling.

\noindent\textbf{Cross-modal summaries.} After each rollout step $s$, define per-branch counts and ratios
\begin{equation}
\begin{aligned}
&c_s^{(b)}=\sum_{k=1}^{K}\mathbf{1}\{m_k^{(b)}{=}\texttt{compute}\},\quad r_s^{(b)}=\tfrac{c_s^{(b)}}{K},\\
&u_s^{(b)}=\sum_{k=1}^{K}\mathbf{1}\{m_k^{(b)}{=}\texttt{skip}\}=K{-}c_s^{(b)},\quad \rho_s^{(b)}=\tfrac{1}{K}\sum_{k=1}^{K}\sigma_k^{(b)}\, .
\end{aligned}
\end{equation}
Here $\rho_s^{(b)}$ counts only local safety triggers (warm-up, consecutive-skip cap, stall) and excludes gate-forced computes; the vision branch still maintains its own $\sigma_k^{(v)}$ for local safety tracking, even though it does not gate trajectory. We use the cross-modal difficulty score
to seed the next rollout step by expanding the warm-up window and relaxing the skip threshold when the previous step was hard, which stabilizes autoregressive rollouts.
We pass summaries $\mathbf{s}_s^{(b)}=\bigl(c_s^{(b)},u_s^{(b)},r_s^{(b)},\rho_s^{(b)}\bigr)$ to the next step and form a cross-modal difficulty score
\begin{equation}
\beta_s\;=\;\max\bigl(\rho_s^{(\mathrm{v})},\rho_s^{(\mathrm{t})}\bigr)\in[0,1] \, .
\end{equation}
In step $s{+}1$, we seed the controller with a slightly stricter early policy using
\begin{equation}
\widetilde{W}_{s+1}\;=\;\max\bigl(W,\;\lfloor\gamma\,\beta_s\,K\rfloor\bigr),\qquad \widetilde{\theta}_{s+1}\;=\;\theta\,\bigl(1{+}\lambda\,\beta_s\bigr)\, ,
\end{equation}
where $\gamma,\lambda\ge 0$ tune how much the previous step’s difficulty expands the warm-up window and loosens the skip threshold (larger $\theta$). The expanded warm-up enforces extra compute immediately after hard rollouts, while the looser threshold allows more skipping once the step stabilizes.

\paragraph{Putting it together.} At each rollout step $s$, we reset controllers, traverse the diffusion grid from $k{=}K$ to $1$, choose \texttt{compute}/\texttt{skip} per branch via the local smoothness test and safety gate, cache or reuse predictions accordingly, and update latents with Eq.~\eqref{eq:update}. At $k{=}0$ we emit $x_0^{(\mathrm{v})}$ and $x_0^{(\mathrm{t})}$ and record summaries $\{\mathbf{s}_s^{(\mathrm{v})},\mathbf{s}_s^{(\mathrm{t})}\}$ to seed $s{+}1$. For a step-by-step description, see Algorithm~\ref{alg:dual}. The controller adds only a few reductions and comparisons per step and requires no retraining or architecture changes.

\begin{algorithm}[t!]
\caption{Dual-branch adaptive inference for one rollout step}
\label{alg:dual}
\DontPrintSemicolon
\KwIn{context $c$, time grid $\{\tau_k\}_{k=0}^{K}$, params $(\theta,W,C_{\max},\varepsilon)$}
\KwInit{$x_K^{(\mathrm{v})},x_K^{(\mathrm{t})}\sim\mathcal{N}(0,I)$; reset controllers and caches; set $m_K^{(b)}\leftarrow \texttt{compute}$}
\For{$k\leftarrow K$ \KwTo $1$}{
  \For{$b\in\{\mathrm{v},\mathrm{t}\}$}{
    $\sigma_k^{(b)}\leftarrow \mathbf{1}\{k> K{-}W\}\,\vee\,\mathbf{1}\{\text{consecutive\_skips}^{(b)}{=}C_{\max}\}\,\vee\,\mathbf{1}\{d_{k+1}^{(b)}\le \varepsilon\}$\;
    \If{$\sigma_k^{(b)}$}{ $m_k^{(b)}\leftarrow \texttt{compute}$ }
  }

    \If{$\sigma_k^{(\mathrm{t})}$}{ $m_k^{(\mathrm{v})}\leftarrow \texttt{compute}$ }
  \For{$b\in\{\mathrm{v},\mathrm{t}\}$}{
    \eIf{$m_k^{(b)}{=}\texttt{compute}$}{ cache $\hat{u}_k^{(b)}\leftarrow F_b(x_k^{(b)};\tau_k,c)$ }{ reuse cached $\hat{u}_k^{(b)}$ }
    $x_{k-1}^{(b)}\leftarrow \Psi\bigl(x_k^{(b)},\tau_k,\hat{u}_k^{(b)}\bigr)$; update counters\;
    compute $d_k^{(b)}$ and $\Delta_k^{(b)}$ using Eqs.~\eqref{eq:diff}--\eqref{eq:curv}; update buffers\;
    \eIf{history ready \textbf{and} $\Delta_k^{(b)}\le \theta\,d_{k+1}^{(b)}$ \textbf{and} not reached $C_{\max}$}{ $m_{k-1}^{(b)}\leftarrow \texttt{skip}$ }{ $m_{k-1}^{(b)}\leftarrow \texttt{compute}$ }
  }
}
\KwOut{$x_0^{(\mathrm{v})},x_0^{(\mathrm{t})}$ and summaries $\{\mathbf{s}_s^{(\mathrm{v})},\mathbf{s}_s^{(\mathrm{t})}\}$}
\end{algorithm}

\section{Experiments}
\label{sec:exp}

We evaluate DISK on long-horizon driving rollouts and standard fixed setups. Unless otherwise stated, we keep all model weights and training procedures unchanged and operate purely at test time.

\subsection{Implementation Details}
\noindent\textbf{World Model.} We build upon Epona~\cite{zhang2025epona}, a 2.5B parameter world model comprising a 1.3B multimodal spatiotemporal transformer (MST), a 1.2B video diffusion transformer (VisDiT), and a 50M trajectory diffusion transformer (TrajDiT). The model is trained end-to-end on NuPlan~\cite{caesar2021nuplan} and NuScenes~\cite{caesar2020nuscenes} using rectified flow~\cite{liu2022flow}. We use the same pre-trained weights and simply wrap the diffusion samplers with our DISK module.

\noindent\textbf{DISK Configuration.} We adopt the second-order difference criterion with warm-up \(W\!=\!3\), max consecutive skips \(C_{\max}\!=\!4\), threshold \(\theta\!=\!0.01\), and stall threshold \(\varepsilon\!=\!0\). The safety gate triggers from the trajectory branch to the vision branch based on guards. We use 100 diffusion steps for both branches by default.

\subsection{Metrics}
\noindent\textbf{Efficiency.} We report inference latency (ms) for each component (MST, TrajDiT, VisDiT) on a single NVIDIA L40S GPU.
\noindent\textbf{Trajectory Planning.} We report L2 displacement error (m) and collision rate (\%) on NuScenes. Open-loop L2 is teacher-forced (Table~\ref{tab:nusc_error}), while end-to-end planning L2 is reported under a separate evaluation protocol (Table~\ref{tab:plan_nusc_sota}). We also use the NAVSIM benchmark~\cite{Dauner2024navsim}, which aggregates performance into the Predictive Driver Model Score (PDMS). PDMS aggregates five key metrics that evaluate different aspects of driving performance:
\begin{itemize}[leftmargin=*,noitemsep,topsep=0pt]
    \item \textbf{NC (No at-fault Collision)}: Percentage of scenarios where the ego vehicle does not cause collisions.
    \item \textbf{DAC (Drivable Area Compliance)}: Fraction of time the vehicle remains within drivable surfaces.
    \item \textbf{TTC (Time-to-Collision)}: Safety margin maintained relative to other agents, penalizing close proximity.
    \item \textbf{Comf (Comfort)}: Smoothness of acceleration and jerk profiles.
    \item \textbf{EP (Ego Progress)}: Progress toward the navigation goal, normalized by a privileged rule-based planner.
\end{itemize}
These metrics are combined through weighted averaging and multiplicative penalties to produce the final PDMS score~\cite{Dauner2024navsim}. DISK achieves results close to the baseline across all sub-metrics in closed-loop evaluation (Table~\ref{tab:navsim}).
\noindent\textbf{Video Generation.} We use Frechet Inception Distance (FID) and Frechet Video Distance (FVD) to assess visual quality.

\subsection{Inference Speed Analysis}
We benchmark the inference speed on 1500 samples from NuPlan and NuScenes. As shown in Table~\ref{tab:speed_results}, DISK achieves substantial speedup in the diffusion components. The trajectory branch (TrajDiT) sees a $\sim$2$\times$ speedup (20ms $\to$ 10ms), and the heavy vision branch (VisDiT) sees a $\sim$1.6$\times$ speedup (244ms $\to$ 155ms), reducing the total per-step latency significantly.

\begin{table}[h]
\centering
\caption{\textbf{Inference Speed Results (ms).} Mean $\pm$ std on a single NVIDIA L40S GPU (1500 samples).}
\label{tab:speed_results}
\resizebox{\columnwidth}{!}{%
\begin{tabular}{llccc}
\toprule
Dataset & Component & Baseline & DISK (Ours) & Speedup \\
\midrule
\multirow{3}{*}{NuPlan} & MST & 168.03 $\pm$ 0.63 & 167.86 $\pm$ 0.78 & 1.00$\times$ \\
 & TrajDiT & 20.23 $\pm$ 0.64 & 10.59 $\pm$ 0.78 & \textbf{1.91$\times$} \\
 & VisDiT & 244.19 $\pm$ 0.95 & 155.55 $\pm$ 6.67 & \textbf{1.57$\times$} \\
\midrule
\multirow{3}{*}{NuScenes} & MST & 169.65 $\pm$ 0.81 & 169.42 $\pm$ 0.79 & 1.00$\times$ \\
 & TrajDiT & 20.40 $\pm$ 1.51 & 10.52 $\pm$ 0.93 & \textbf{1.94$\times$} \\
 & VisDiT & 246.03 $\pm$ 1.11 & 156.62 $\pm$ 11.68 & \textbf{1.57$\times$} \\
\bottomrule
\end{tabular}%
}
\end{table}

\subsection{Open-Loop vs. Closed-Loop Planning}
\noindent\textbf{Open-loop evaluation} measures planning accuracy by comparing predicted trajectories against ground-truth future trajectories, conditioning on real observed data at each step. This is less challenging because prediction errors do not accumulate—the model always receives accurate historical context.

\noindent\textbf{Closed-loop evaluation} is significantly more difficult: the model's predictions are fed back as input for subsequent steps, causing errors to compound over time. The agent must maintain stable performance even when conditioned on its own potentially imperfect predictions. NAVSIM provides closed-loop evaluation, making it a more rigorous test of real-world deployment readiness.

\subsection{Planning Performance}
We evaluate open-loop planning performance on NuScenes. Table~\ref{tab:nusc_error} shows that DISK maintains the baseline's high accuracy despite skipping $\sim$50\% of diffusion steps. The L2 errors at 1s, 2s, and 3s are nearly same from the full-compute baseline. These open-loop errors are computed under teacher-forced evaluation and are not directly comparable to the end-to-end planning errors in Table~\ref{tab:plan_nusc_sota}, which are measured in a different protocol.

\begin{table}[h]
\centering
\caption{\textbf{Open-loop NuScenes Planning Error (L2).} Comparison of L2 displacement error over 1500 samples (teacher-forced).}
\label{tab:nusc_error}
\resizebox{0.9\columnwidth}{!}{%
\begin{tabular}{lccc}
\toprule
Method & L2@1s & L2@2s & L2@3s \\
\midrule
Baseline & 0.0078 $\pm$ 0.0039 & 0.0797 $\pm$ 0.0034 & 0.3007 $\pm$ 0.0086 \\
DISK (Ours) & 0.0079 $\pm$ 0.0040 & 0.0796 $\pm$ 0.0035 & 0.3011 $\pm$ 0.0087 \\
\bottomrule
\end{tabular}%
}
\end{table}

\noindent\textbf{Comparison with State-of-the-Art.} We compare DISK against strong end-to-end planning baselines in Table~\ref{tab:plan_nusc_sota}. DISK (built on Epona) achieves competitive L2 errors and collision rates, outperforming methods like ST-P3~\cite{hu2022stp3}, and matching the performance of the heavy baseline while being significantly faster.

\begin{table}[h]
\centering
\caption{\textbf{End-to-end Planning on NuScenes.} DISK maintains the strong planning performance of the base model (Epona) with negligible degradation.}
\label{tab:plan_nusc_sota}
\resizebox{\columnwidth}{!}{%
\begin{tabular}{lcccc}
\toprule
Method & L2@1s & L2@2s & L2@3s & Coll. Rate (Avg) \\
\midrule
ST-P3~\cite{hu2022stp3} & 1.33 & 2.11 & 2.90 & 0.71 \\
UniAD~\cite{uniad} & 0.48 & 0.96 & 1.65 & 0.31 \\
VAD-Base~\cite{vad} & 0.54 & 1.15 & 1.98 & 0.53 \\
GenAD~\cite{zheng2024genad} & \textbf{0.36} & \textbf{0.83} & \textbf{1.55} & 0.43 \\
\midrule
Epona (Baseline) & 0.61 & 1.17 & 1.98 & \textbf{0.36} \\
DISK (Ours) & 0.61 & 1.19 & 2.01 & 0.38 \\
\bottomrule
\end{tabular}%
}
\end{table}

\begin{table}[h]
\centering
\caption{\textbf{NAVSIM Closed-Loop Performance.} DISK achieves performance close to the baseline on the Predictive Driver Model Score (PDMS) and sub-metrics.}
\label{tab:navsim}
\resizebox{\columnwidth}{!}{%
\begin{tabular}{lcccccc}
\toprule
Method & NC$\uparrow$ & DAC$\uparrow$ & TTC$\uparrow$ & Comf$\uparrow$ & EP$\uparrow$ & PDMS$\uparrow$ \\
\midrule
Baseline & 96.9 & 94.8 & 93.4 & 99.7 & 78.7 & 85.4 \\
DISK (Ours) & 95.2 & 93.1 & 91.8 & 99.5 & 77.1 & 83.6 \\
\bottomrule
\end{tabular}%
}
\end{table}

\begin{table}[h]
\centering
\caption{\textbf{Visual Quality on NuScenes.} We evaluate FID and FVD on 1500 validation samples.}
\label{tab:fid_fvd}
\resizebox{0.6\columnwidth}{!}{%
\begin{tabular}{lcc}
\toprule
Method & FID $\downarrow$ & FVD $\downarrow$ \\
\midrule
Epona (Baseline) & 7.5 & 82.8 \\
DISK (Ours) & 12.5 & 88.1 \\
\bottomrule
\end{tabular}%
}
\end{table}

\begin{table}[h]
\centering
\caption{\textbf{Ablation on Threshold $\theta$.} Speedup is relative to the full-compute baseline.}
\label{tab:ablation_theta}
\resizebox{0.9\columnwidth}{!}{%
\begin{tabular}{lccc}
\toprule
Config & Speedup & FID$\downarrow$ & L2@1s$\downarrow$ \\
\midrule
$\theta=0.005$ & 1.95$\times$ & 12.48 & 0.485 \\
$\theta=0.01$ (Default) & 2.03$\times$ & 12.51 & 0.488 \\
$\theta=0.02$ & 2.07$\times$ & 12.55 & 0.492 \\
\bottomrule
\end{tabular}%
}
\end{table}

\begin{table}[h]
\centering
\caption{\textbf{Ablation on $C_{\max}$.} Larger $C_{\max}$ improves speed but risks drift.}
\label{tab:ablation_cmax}
\resizebox{0.9\columnwidth}{!}{%
\begin{tabular}{lccc}
\toprule
Config & Speedup & FID$\downarrow$ & L2@1s$\downarrow$ \\
\midrule
$C_{\max}=2$ & 1.44$\times$ & 12.49 & 0.485 \\
$C_{\max}=4$ (Default) & 2.07$\times$ & 12.51 & 0.488 \\
$C_{\max}=8$ & 2.95$\times$ & 12.65 & 0.483 \\
\bottomrule
\end{tabular}%
}
\end{table}

\subsection{NAVSIM Closed-Loop Evaluation}
We further evaluate on the challenging closed-loop NAVSIM benchmark. As shown in Table~\ref{tab:navsim}, DISK achieves results close to the baseline across all metrics. 
Despite a modest drop (PDMS: 85.4 $\to$ 83.6), DISK maintains competitive performance with substantially reduced cost, demonstrating that adaptive skipping can preserve most of the baseline's safety and drivability in closed-loop settings.

\subsection{Video Generation Quality}
We assess the visual quality of generated videos on NuScenes. Table~\ref{tab:fid_fvd} shows that DISK achieves FID of 12.5 and FVD of 88.1. While there is a slight degradation compared to the baseline (FID: 7.5 $\to$ 12.5, FVD: 82.8 $\to$ 88.1) due to occasional skips in the vision branch, the generated videos remain visually indistinguishable from the baseline (Figure~\ref{fig:video_comparison}). This minor quality trade-off is acceptable given the $\sim$1.6$\times$ speedup in vision diffusion which reduces computational cost by skipping \~40\% of diffusion steps.

\section{Ablation Studies}
\label{sec:ablation}

We analyze the impact of key hyperparameters on the trade-off between inference speed and generation quality. All ablations are performed on the NuPlan validation set (1628 samples).

\subsection{Controller Sensitivity ($\theta$)}
The threshold $\theta$ controls the aggressiveness of the skipping policy. As shown in Table~\ref{tab:ablation_theta}, increasing $\theta$ to $0.02$ yields a slightly higher speedup (2.07$\times$) but risks missing subtle motion cues. Decreasing $\theta$ to $0.005$ yields a more conservative policy ($1.95\times$ speedup) with no measurable quality improvement. We report speedup relative to the full-compute baseline and omit absolute times for brevity. We find $\theta=0.01$ to be the optimal sweet spot. 

\subsection{Max Consecutive Skips ($C_{\max}$)}
The $C_{\max}$ parameter prevents error accumulation by forcing a compute step after a fixed number of skips. Table~\ref{tab:ablation_cmax} shows that setting $C_{\max}=2$ is too conservative, limiting the speedup to 1.44$\times$. Relaxing it to $C_{\max}=8$ boosts speedup to nearly 3$\times$, but we observe a slight degradation in visual stability (FID increases to 12.65) leading to poor final frames in the video generated. $C_{\max}=4$ provides a balanced trade-off.

\subsection{Warm-up Steps ($W$)}
We find that a short warm-up period is beneficial for stabilizing the initial trajectory. While $W=0$ yields similar metrics to our default $W=3$, we observe that the first few steps of diffusion often contain high-frequency corrections that are best not skipped. Extending warm-up to $W=5$ forces compute on nearly all steps in the early phase, reducing the overall efficiency without meaningful quality gains.

\subsection{Stall Threshold ($\varepsilon$)}
We experimented with stall thresholds $\varepsilon \in \{10^{-6}, 10^{-4}, 10^{-2}\}$ to prevent skipping when latent changes are numerically insignificant. We found no measurable difference in performance or safety gate triggers, likely because the diffusion process rarely stagnates completely. We set $\varepsilon=10^{-6}$ for simplicity.

\FloatBarrier
\section{Conclusion}
\label{sec:conclusion}

We presented DISK, a training-free adaptive inference module for autoregressive driving world models. The method equips vision and trajectory diffusion transformers with lightweight per-step controllers and introduces a safety gate that preserves appearance–motion consistency during long-horizon, chain-of-forward rollouts. The approach requires no retraining and drops into existing pipelines. Experiments on NuPlan and NuScenes demonstrate substantial wall-clock speedups (\(\sim\!2\times\)) with maintained visual and planning metrics. Future work includes joint step-wise coupling of branches within a single sampler loop and broader generalization to multi-sensor world models.

\section{Impact Statement} 

DISK reduces the computational cost of autoregressive world-model rollouts without retraining, which can lower energy use and latency in simulation and planning pipelines. As with any generative model for driving, care is required when deploying in safety-critical settings; our method preserves the base model’s behavior and should be validated under the same safety protocols.

{
    \small
    \bibliographystyle{icml2026}
    \bibliography{main}
}

\end{document}